# Mixed-Variable Global Sensitivity Analysis For Knowledge Discovery And Efficient Combinatorial Materials Design

**Yigitcan Comlek**
Department of Mechanical Engineering, Northwestern University
2145 Sheridan Road, Evanston, IL 60208
yigitcancomlek2024@u.northwestern.edu

**Liwei Wang**
Department of Mechanical Engineering, Northwestern University
2145 Sheridan Road, Evanston, IL 60208
liwei.wang@northwestern.edu

**Wei Chen**[*]
Department of Mechanical Engineering, Northwestern University
2145 Sheridan Road, Evanston, IL 60208
weichen@northwestern.edu

[*] Corresponding Author

**ABSTRACT**

*Global Sensitivity Analysis (GSA) is the study of the influence of any given inputs on the outputs of a model. In the context of engineering design, GSA has been widely used to understand both individual and collective contributions of design variables on the design objectives. So far, global sensitivity studies have often been limited to design spaces with only quantitative (numerical) design variables. However, many engineering systems also contain, if not only, qualitative (categorical) design variables in addition to quantitative design variables. In this paper, we integrate Latent Variable Gaussian Process (LVGP) with Sobol' analysis to develop the first metamodel-based mixed-variable GSA method. Through numerical case studies, we validate and demonstrate the effectiveness of our proposed method for mixed-variable problems. Furthermore, while the proposed GSA method is general enough to benefit various engineering design applications, we integrate it with multi-objective Bayesian optimization (BO) to create a sensitivity-aware design framework in accelerating the Pareto front design exploration for metal-organic framework (MOF) materials with many-level combinatorial design spaces. Although MOFs are constructed only from qualitative variables that are notoriously difficult to design, our method can utilize sensitivity analysis to navigate the optimization in the many-level large combinatorial design space, greatly expediting the exploration of novel MOF candidates.*

## 1. INTRODUCTION

In the context of engineering design research, sensitivity analysis (SA) can be defined as the study of how the input design variables, independently or interactively, influence the outputs, i.e., the design objectives of interest [1]. SA plays a crucial role in aiding designers in constructing effective design models and processes by identifying design factors that contribute most to the goals, calibrating model parameters, quantifying input and output uncertainties, reducing design space, and verifying whether the overall behavior of a model matches with the real system.

SA can be considered on two levels: local and global [2]. Local sensitivity analysis (LSA) focuses on understanding the impact of local perturbations of design variables on the design objective [3]. Specifically, in most cases, LSA is carried out through calculating the partial derivatives of the design objective with respect to the design variables. As a result, LSA is considered to be a one-factor-at-a-time approach as only one design variable is varied at a time while the other variables are held constant [4]. Although local SA can be beneficial for gradient-based design exploration and optimization, it fails to provide a more comprehensive understanding of design variables on the response, specifically when the design system of interest is nonlinear and various design variables are impacted by the system of interest simultaneously [5]. On the other hand, global sensitivity analysis (GSA) is a comprehensive method that assesses the influence of the design variables on the design objective over the entire design space. As a result, the main goal of GSA is to reveal the importance and interactions of design variables on the design objective at a global level to systematically quantify and improve the understanding of the design system. The GSA method possesses two distinctive properties. First, the sensitivity results obtained from global methods incorporate the influence of the entire ranges and the probability density distributions of the design variables. Second, the sensitivity results of each design variables are calculated while varying other design variables simultaneously [4]. Among various GSA approaches, variance-based GSA is the most commonly used in many engineering applications [6-8]. Variance-based methods aim to analyze how much of the overall variability of the response can be related to the variability of the design variables [9]. Within the variance-based methods, Sobol' indices prevail as the sensitivity measures that explain the contribution of each design variable on the variability of the response through two sensitivity metrics, i.e., Main Sensitivity Index (MSI) and Total Sensitivity Index (TSI) [10]. The MSI aims to quantify how each design variable contributes to the response individually, whereas the TSI quantifies the interaction and collective effect within design variables on the response. Evaluations of Sobol' metrics typically require thousands of samples for statistical analysis, which is infeasible if the design objective is computationally expensive to obtain [11, 12]. However, with the recent advancements in metamodeling and machine learning, the issue of high computational cost in evaluating responses has been mitigated by using efficient and yet accurate metamodels, including but not limited to Kriging models, Neural Networks, and Gaussian Process models. Within this context, Chen et al. have developed analytical sensitivity analysis using metamodels to further accelerate the calculations of Sobol' sensitivity metrics [13].

Thus far, to the best of the authors' knowledge, the GSA techniques are limited to studies with only quantitative (numerical) design variables, even though qualitative (categorical) design variables are ubiquitous in many engineering applications such as materials design and topology optimization [14, 15]. There is a clear need for incorporating qualitative design variables into GSA studies to understand their influences on the design objectives. The key challenge hindering the application of GSA for qualitative variables is the lack of a corresponding sampling technique and accurate surrogate modeling for mixed-variable design problems. The aim of this research is thus to create a mixed-variable GSA method by integrating Latent Variable Gaussian Process (LVGP) machine learning model and a new mixed-variable sampling methodology for metamodel-based GSA studies. The LVGP allows the incorporation of qualitative variables into Gaussian Process (GP) models by implicitly mapping each qualitative variable into a quantitative space through low-dimensional latent variables [16]. With this approach, the qualitative variables can be directly implemented into the GSA studies through metamodel-based sensitivity evaluations. In this study, we demonstrate our method using Sobol' indices, as it is one of the most commonly used GSA methods in the design community due to its advantages in handling high-dimensional design variables, capturing non-linear responses, and providing quantitative sensitivity measures of both individual and interaction of design variables [10, 17]. However, we emphasize that the two key components of the proposed approach, i.e., the mixed-variable metamodeling with LVGP and sampling strategy, are readily applicable to any metamodeling-based GSA methods. Furthermore, the mixed-variable GSA approach is integrated with Bayesian optimization, utilizing the sensitivity information to accelerate the search in many-level combinatorial materials design spaces.

The outline of the paper is as follows. In Section 2, a brief introduction of the LVGP and metamodel-based Sobol' sensitivity indices is given. Next, the proposed mixed-variable GSA method is explained in detail. In Section 3, the method is tested on two mathematical functions commonly used in literature for GSA studies to validate its efficacy. Finally, a design framework integrating mixed-variable GSA and Bayesian optimization is demonstrated to accelerate the design of novel metal-organic framework (MOF) materials. The mixed-variable GSA method has been shown to accelerate the exploration of optimum solutions in a large combinatorial design space.

## 2. METAMODEL-BASED MIXED-VARIABLE GLOBAL SENSITIVITY ANALYSIS METHOD

With the integration of LVGP and metamodel-based Sobol' indices, we propose metamodel-based mixed-variable GSA method to incorporate qualitative variables into GSA applications. In this section, we give brief reviews of LVGP and metamodel-based GSA methods for Sobol' indices calculations. Next, we demonstrate the integration of LVGP for Sobol' indices calculations and provide the details of our proposed mixed-variable GSA method.

### 2.1. Review of Latent Variable Gaussian Process (LVGP)

Due to their strengths in delivering fast surrogate modeling, capturing nonlinear responses, and providing objective and uncertainty predictions, Gaussian processes (GPs) are well-known metamodels that have been used in many engineering applications [18]. However, due to the nature of the GP correlation functions, qualitative (categorical) variables cannot be implemented directly without a well-defined distance metric. To overcome this challenge, Latent Variable Gaussian Process (LVGP) has been developed to incorporate qualitative variables into GP modeling through quantitative latent variable representation of these variables [16]. The key idea behind the LVGP modeling is that for every qualitative variable, there exists an underlying, potentially high-dimensional, quantitative space that explains the qualitative variable's influence on the response of interest. Using this knowledge, LVGP aims to construct an implicit mapping from qualitative variables into a low-dimensional quantitative latent space, approximating the effects of underlying variables on the response.

To make the demonstration of LVGP more concrete, consider a design space with mixed-variable input, $\boldsymbol{w} = [\boldsymbol{x}^{\boldsymbol{T}}, \boldsymbol{t}^{\boldsymbol{T}}]^T$ where $\boldsymbol{x} = [x_1, x_2, \ldots, x_q]^T \in R^q$ are quantitative variables and $\boldsymbol{t} = [t_1, t_2, \ldots, t_m]^T$ are qualitative (categorical) design variables. Each qualitative design variable $t_j$ has $l_j$ design options (levels), i.e., $t_j \in \{1,2, \ldots, l_j\}l$ for $j = 1,2, \ldots m$. Here, it is highly important to note that the integers ranging from $1,2, \ldots, l$ are *only* used as indexes for labeling different design choices and are not used in the Gaussian Process kernel training. An example of qualitative variables can be the elements that are used in the design of MOFs (e.g., Cu, Co, Ni, Zn options for the organic (nodular) building block, $l_j = 4$ ). For real physical models, there are quantitative variables $\boldsymbol{v}(t_j) = [v_1(t_j), v_2(t_j) \ldots, v_n(t_j)] \epsilon \mathrm{R}^n$ underlying each qualitative variable, which could be unknown, undiscovered, or extremely high-dimensional. The main idea of LVGP is to learn a low-dimensional latent space to approximate the original underlying qualitative space via statistical inference. Although the dimensions of the learned latent variable vector $\boldsymbol{z} \epsilon \mathrm{R}^k$ can be freely chosen, a two-dimensional (2D) latent vector, $k = 2$ , is usually sufficient in most engineering designs [16], which is also adopted in this study. Therefore, each level of a qualitative variable $t_j$ is represented by a 2D latent vector $\boldsymbol{z}(t_j) = [z_{j,1}, z_{j,2}]^T$. Then, the transformed design space becomes $\boldsymbol{h} = [\boldsymbol{x}^{\boldsymbol{T}}, \ \boldsymbol{z}(\boldsymbol{t})^T]^T \in R^{(q+m\times 2)}$ with $\boldsymbol{z}(\boldsymbol{t}) =$ $\left[z_{1,1}, z_{1,2}, \dots, z_{j,1}, z_{j,2}, \dots, z_{m,1}, z_{m,2}\right]^T$. An illustration of the implicit mapping from qualitative variables, represented by different shapes, to two-dimensional latent variables is shown in Figure 1.

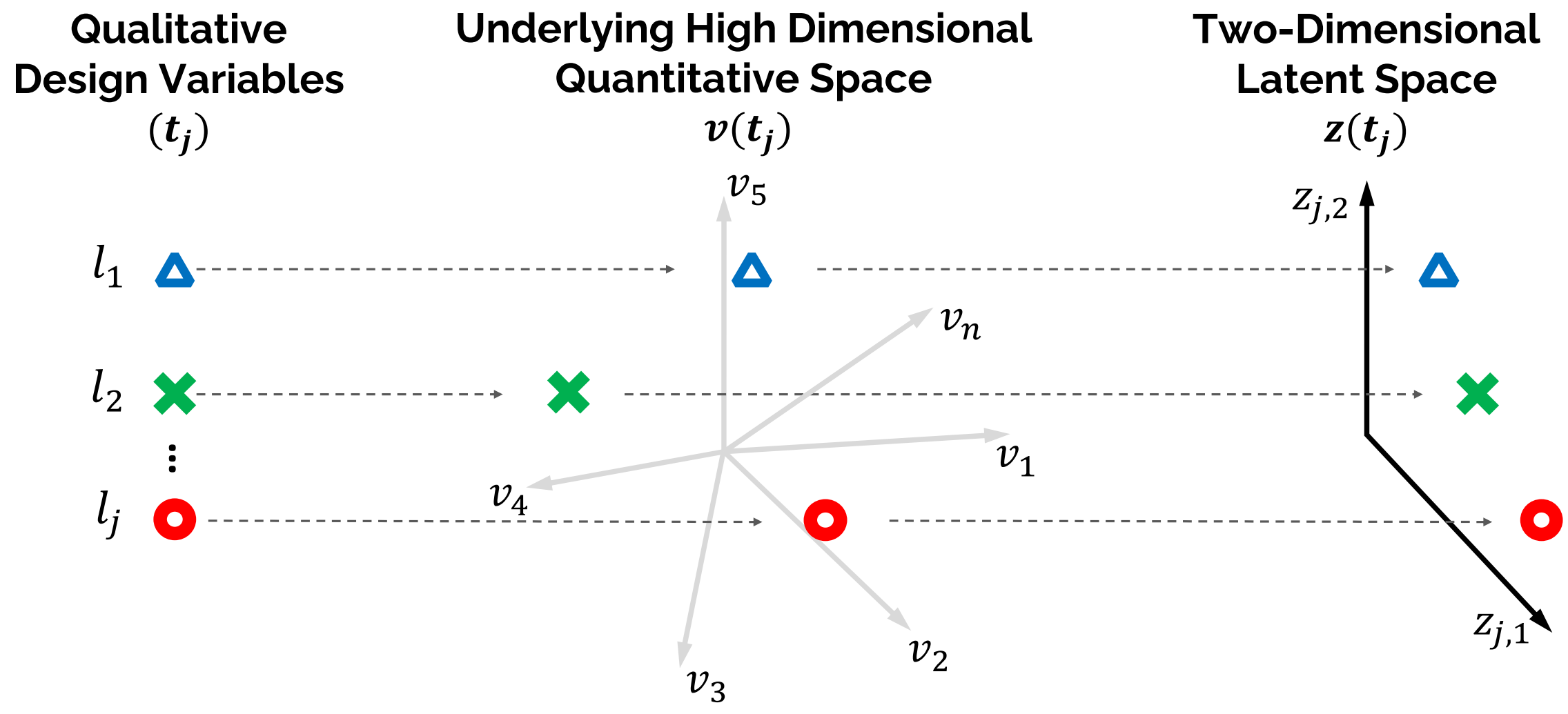

**FIGURE 1**: The mapping from qualitative design variables onto 2-dimensional quantitative latent space

Now, consider a single response GP model with a prior constant mean $\mu$ to describe the mean response at any given point in the transformed design space $\boldsymbol{h}$. A zero-mean Gaussian Process is used to capture the variance of the response, described by a covariance function $K(\boldsymbol{h}, \boldsymbol{h}')$. The covariance function $K(\boldsymbol{h}, \boldsymbol{h}') = \sigma^2 \cdot c(\boldsymbol{h}, \boldsymbol{h}')$ describes the relationship or the correlation of responses at any pairs of input points $\boldsymbol{h}$ and $\boldsymbol{h}'$, where $\sigma^2$ represents the prior variance of the GP model and $c(\boldsymbol{h}, \boldsymbol{h}')$ is the correlation function. LVGP extends the commonly used Gaussian correlation function to include latent variables [16],

$$c(\boldsymbol{h}, \boldsymbol{h}') = exp\left(\sum_{i=1}^{q} \phi_i (x_i - x_i')^2 - \sum_{j=1}^{m} \left\| z_{j,1} - z_{j,1}' \right\|_2^2 + \left\| z_{j,2} - z_{j,2}' \right\|_2^2 \right), \quad (1)$$

where $\phi_i$ is a scaling parameter that will be estimated for each quantitative variable $x_i$. Note that the scaling parameters of latent variables are a unity vector since these scaling factors are absorbed into the estimated latent variable values $\boldsymbol{z}(\boldsymbol{t})$. The rationale behind this correlation function is that points that are spatially closer in the design space $\boldsymbol{h}$ should also exhibit similar output patterns. For a size-$n$ training sample set with an input $\boldsymbol{W_s} = \left[\boldsymbol{w}^{(1)}, \boldsymbol{w}^{(2)}, \dots, \boldsymbol{w}^{(n)}\right]^T$ and an output $\boldsymbol{y_s} = \left[y^{(1)}, y^{(2)}, \dots, y^{(n)}\right]^T$, the parameters, $\mu$, $\sigma$, and $\boldsymbol{\phi} = [\phi_1, \phi_2, \dots, \phi_q]$, along with the 2D mapped latent variables, $\boldsymbol{z}(\boldsymbol{t})$, are estimated through Maximum Likelihood Estimation (MLE) [16], i.e., finding parameters to maximize the log-likelihood function,

$$l(\mu, \sigma, \boldsymbol{\phi}, \boldsymbol{z}) = -\frac{n}{2} ln(\sigma^2) - \frac{1}{2} ln|\boldsymbol{C}(\boldsymbol{z}, \boldsymbol{\phi})| - \frac{1}{2\sigma^2} (\boldsymbol{y_s} - \mu \mathbf{1})^T \boldsymbol{C}(\boldsymbol{z}, \boldsymbol{\phi})^{-1} (\boldsymbol{y_s} - \mu \mathbf{1}), \quad (2)$$

where $\boldsymbol{C}$ is the $n \times n$ correlation matrix with $C_{ij} = c(\boldsymbol{h}^{(i)}, \boldsymbol{h}^{(j)})$ for $i, j = 1,2, \dots, n$, $\boldsymbol{h}^{(i)}$ is the transformed input from the sample $\boldsymbol{w}^{(i)}$, $\mathbf{1}$ is a vector of ones with dimensions of $n \times 1$. Once the latent variables $\boldsymbol{z}(\boldsymbol{t})$ and the parameters $\mu, \sigma$, and $\boldsymbol{\phi}$ are estimated, the fitted model will be used for metamodel-based mixed-variable GSA. Readers interested further in LVGP are referred to the original paper [16].

### 2.2. Metamodel-Based Sobol' Indices Calculations for Global Sensitivity Analysis

Sobol' sensitivity indices are prevailing measures of variation-based GSA, with the Main Sensitivity Index (MSI) and Total Sensitivity Index (TSI) indicating the contribution of each design variable and their interactions on the response's variation, respectively. For a design system with $q$ continuous design variables, the Sobol' indices can be calculated as below:

$$S_i = \frac{Var_x\left[E_{\boldsymbol{x}_1,\dots,\boldsymbol{x}_q} y \middle| \boldsymbol{x}_i\right]}{Var_{\boldsymbol{x}_1,\dots,\boldsymbol{x}_q}[y]} = \frac{V_i}{V}, for\ i = 1, \dots, q, \tag{3}$$

$$S_i^t = S_i + S_{i,\sim i},\ \ for\ i = 1, \dots, q, \tag{4}$$

where $\boldsymbol{y}$ is the response of interest, $V_i$ is the variance of the response with respect to the changes in design variable $x_i$, $V$ is the variance of the response, $S_i$ is the MSI, $S_i^t$ is the TSI of the design variables, and $S_{i,\sim i}$ is the higher order Sobol' sensitivity indices between variable $x_i$ and remaining variables $x_{\sim i}$. To evaluate the indices, Monte Carlo sampling approach is usually adopted, requiring a large number of response evaluations that are unaffordable for design applications with high costs. To overcome this challenge, Chen et al. developed metamodel-based Sobol' indices calculations through tensor product formulation [13]. This methodology enabled the integration of metamodels to provide fast and accurate response evaluation for GSA studies. Herein, we use LVGP as our mixed-variable metamodel and extend the formulation in [13] to be

$$S_i = \frac{Var_{w_i} E_{w,\dots,w_d}\left[E\left[y_{LVGP|\boldsymbol{W}_s,\boldsymbol{y}_s}(w)\right] \middle| w_i\right]}{Var_{w_i,\dots,w_d} E\left[y_{LVGP|\boldsymbol{W}_s,\boldsymbol{y}_s}(w)\right]}, for\ i = 1, \dots, d, \tag{5}$$

Here, $\boldsymbol{w} = [\boldsymbol{x}^{\boldsymbol{T}}, \boldsymbol{t}^{\boldsymbol{T}}]^T \in \mathrm{R}^d$ with $d = q + m$ corresponds to the original mixed-variable design space with quantitative $(x_q)$ and qualitative $(t_m)$ variables, and $y_{LVGP|W_s,y_s}$ is the predicted response using the LVGP model trained on the sample set $\boldsymbol{W_s}$, and response $\boldsymbol{y_s}$. It should be noted that the 2D latent variables are employed in the correlation function for response prediction only, while Sobol' indices are calculated by directly sampling mixed-variable design space, using the original qualitative variable $\boldsymbol{t}$ instead of its 2D latent variables. More details on the sampling are described in the next section. Although there are other metamodel options that can also handle mixed-variable inputs, such as Random Forest (RF), the prediction accuracy of LVGP was shown to be better in most cases given the same sample size [16]. Therefore, we chose LVGP as our metamodel in this study. With the integration of metamodel-based GSA calculations and LVGP, we can then incorporate qualitative variables into GSA applications.

### 2.3. Metamodel-Based Mixed-Variable Global Sensitivity Analysis Method

Given a mixed-variable design space, the GSA is realized through the method shown in Figure 2. First, the mixed-variable samples are fed into the LVGP for model fitting. One of the most important factors in obtaining accurate sensitivity indices relies on the sampling strategy as it determines the coverage of the design space. A well-designed sampling strategy should aim to evenly sample the entire range of the design space to ensure that the important regions of the design space are covered. Once the trained model is ready, the Sobol' analysis starts with sampling the qualitative and quantitative variables from quasi-random Sobol sets, filling the given mixed-variable design space range in a highly uniform manner. In particular, to achieve discrete sampling for qualitative variables, once the Sobol set is created, the sample set is sliced into $l_j$ equal sections for each qualitative variable $t_j$. The corresponding level can be obtained based on which section a specific sample falls into. In this way, each level has an equal chance to be sampled from the quasi-random set. Next, the LVGP model is employed to make fast and accurate predictions on the sample space. Finally, using the tensor product formulation given in Equation (5), the Sobol' sensitivity metrics, MSI and TSI are obtained. It is important to note that the amount of training data can influence the accuracy of the LVGP model, which in return determines the accuracy of the sensitivity indices, as in other GSA methods. Therefore, it is important to have enough training samples to train the LVGP metamodel. Although the demonstrated framework is named for mixed-variable design spaces, it is also applicable for design spaces with only qualitative variables as we will demonstrate later with applications to design of MOF materials.

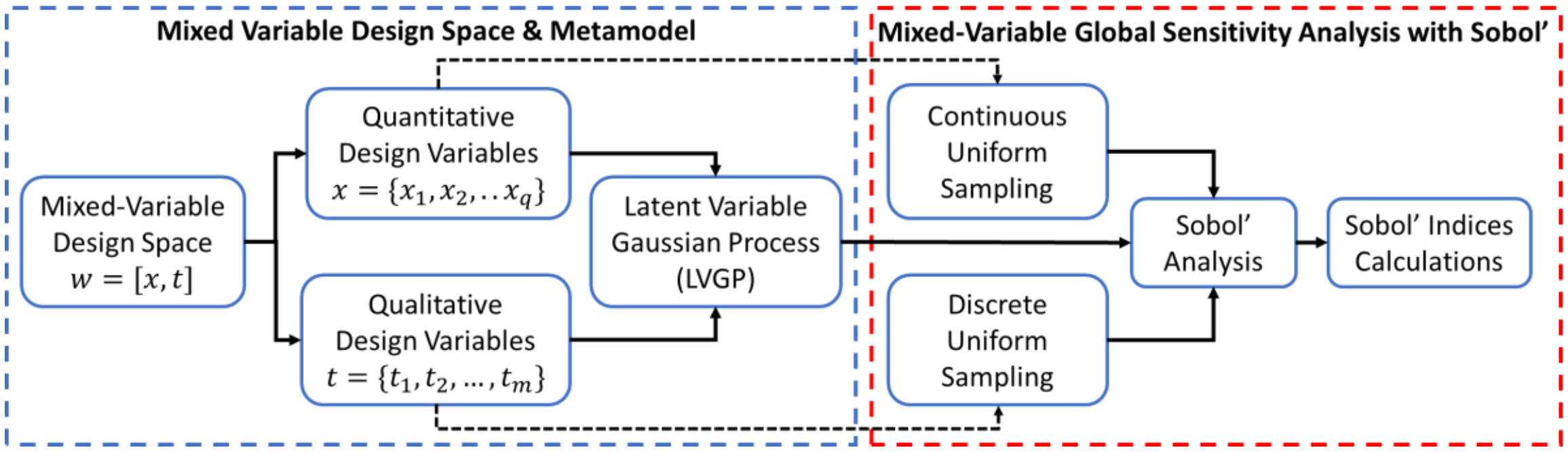

**FIGURE 2**: The flowchart of the metamodel-based mixed-variable global sensitivity analysis method

## 3. VALIDATION WITH NUMERICAL EXAMPLES

Before applying our method to real engineering design applications, we validated it through two well-known mathematical testing functions for GSA studies: Ishigami and Hartmann 6D functions [19]. Since these test functions only have quantitative variables in their original forms, we first converted some design variables into qualitative variables by assigning some predetermined discrete values to the variables. Once again, it is important to note here that the discrete values are treated levels to the LVGP model and are not used in the LVGP model training. This allows us to convert quantitative design spaces into mixed-variable design spaces as test cases for our mixed-variable GSA. As the design spaces are converted to mixed-variable sets, we expect that the influence of both the existing quantitative variables and the newly generated qualitative variables will be different from their effect for test functions of their original forms. Therefore, we implemented a two-stage verification to validate our mixed-variable GSA method. First, we compared the Mixed-Variable (MV) Sobol' indices obtained from our method with the Ground-Truth Mixed-Variable (True-MV) Sobol' indices. Here the True-MV indices correspond to the ground-truth value of Sobol' metrics when the design space is mixed-variable. Since the testing functions are known beforehand, the true values of the metrics can be obtained through the sampling scheme defined in Section 2.3. The validation at this stage checks whether the mixed-variable metrics obtained from our method are matching well with the true mixed-variable metrics. In the second stage, we compared the MV Sobol' indices with the ground-truth (True) Sobol' values obtained from the GSA with only quantitative design variables, i.e., in their original forms. The key idea behind this validation is that as the number of levels, i.e., design choices, increase, the Sobol' indices values must approach the true Sobol' metrics with only quantitative variables. This is due to the fact that the continuous sampling space can be considered as a discrete space with an infinite number of levels. As a result, we expect to see the convergence of MV Sobol' indices towards the ground-truth Sobol' indices values as the number of levels increases. With the defined two validation criteria, we moved on to validate our method on the two mathematical functions.

### 3.1. Validation with the Ishigami Function

The Ishigami function is a highly non-linear testing function with three variables that have been extensively used in GSA studies [20]. The formulation of the Ishigami function, as well as its ground-truth Sobol' sensitivity indices, are given in Equation (6) and Table 1, respectively.

$$f(\boldsymbol{x}) = \sin(x_1) + 7\sin^2(x_2) + 0.1x_3^4\sin(x_1),\ for\ \boldsymbol{x} = [x_1, x_2, x_3]\ \epsilon\ [-\pi, \pi], \quad (6)$$

**TABLE 1.** Ground-truth Sobol' sensitivity indices of the original Ishigami function with only quantitative inputs

| Design Variables | Main Sensitivity Index (MSI) | Total Sensitivity Index (TSI) |
|---|---|---|
| $x_1$ | 0.3138 | 0.5575 |
| $x_2$ | 0.4413 | 0.4424 |
| $x_3$ | 0 | 0.2436 |

To implement the Ishigami function as a mixed-variable problem, we converted the design variables $x_1$ and $x_3$ to qualitative variables by assigning different discrete values and treating them as levels between the given ranges in Equation (6). For ease of illustration, we renamed $x_1$ as $t_1$ and $x_3$ as $t_3$. For instance, in the case of 3 levels, the variables $t_1$ and $t_3$ can only take the levels of $(-\pi, 0, \pi)$. To test each validation stage, we looked at cases with a number of levels ranging from $[2, 20]$. For each level instance, we trained an LVGP model to learn the response surface of the Ishigami function. Next, we used our proposed method to calculate the MSI and TSI for all instances. Finally, we obtained the ground-truth discrete (True-MV) and quantitative (True) Sobol' metrics from the Ishigami function and compared the results. Figure 3 shows both the MSI and TSI values obtained from our validation study. In the figure, the first column shows the MSI, and the second column shows the TSI for all three design variables.

For the cases with qualitative variables, $t_1$ and $t_3$, we observe a great match between the true discrete Sobol' values (True-MV MSI & TSI) and the values obtained from our method (MV MSI & TSI). Specifically, the method was able to identify not only the zero individual contribution provided by $t_3$, but also its total interaction contribution with other design variables. Furthermore, although $x_2$ was considered as a quantitative variable, we expected to observe a notable change in its Sobol' values as the number of levels associated with the other two qualitative variables vary. Therefore, we examined the changes in Sobol' values of $x_2$ as a function of the number of levels of $t_1$ and $t_2$. Our method was able to capture this change in Sobol' values very well for both the MSI and TSI. Finally, the convergence of Sobol' metrics to the true quantitative Sobol' sensitivity values (True TSI & MSI) demonstrated and validated the correct implementation of our method.

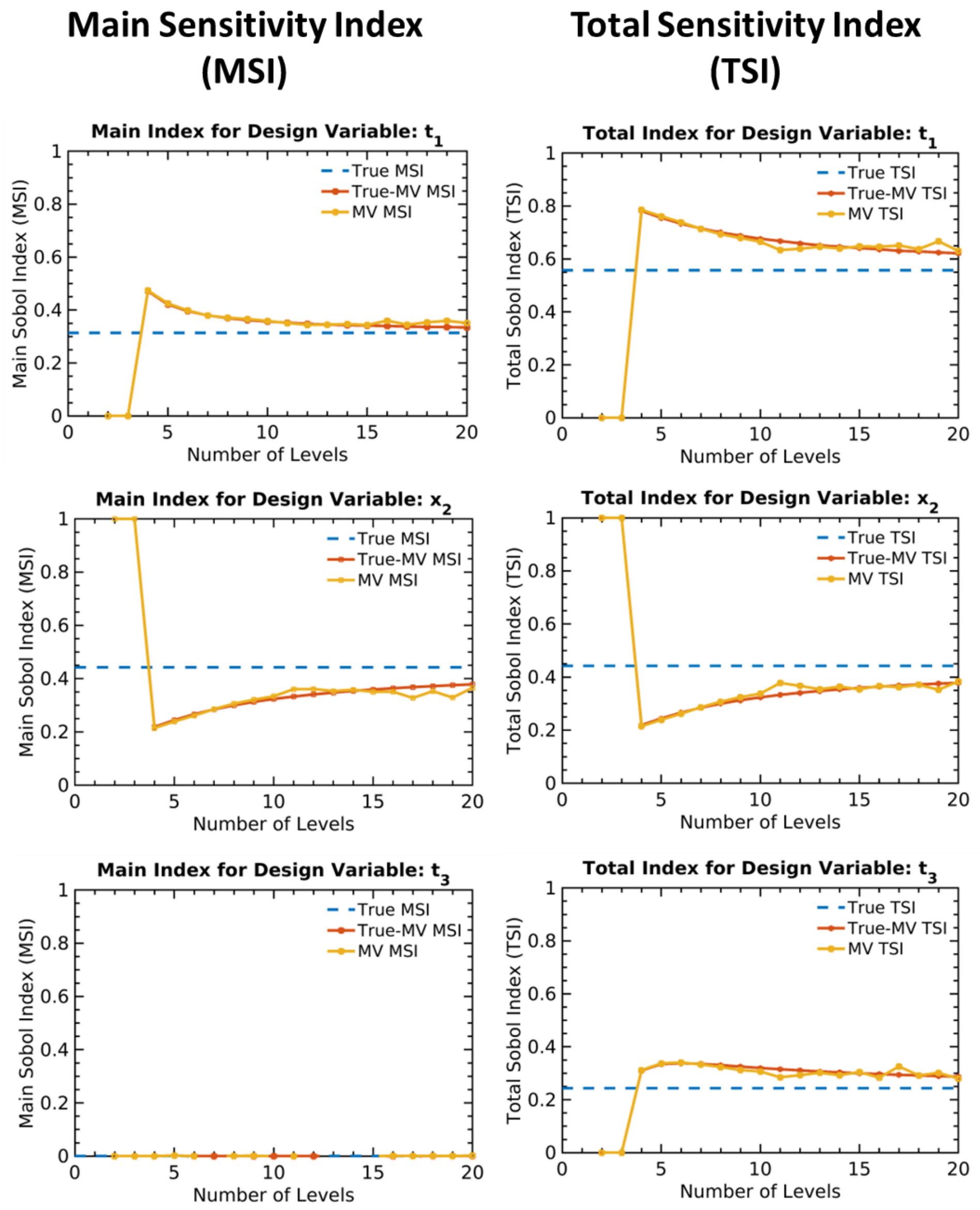

**FIGURE 3**: Sobol' sensitivity indices of the modified Ishigami function with mixed-variable inputs

### 3.2. Validation with the Hartmann 6D Function

We continued to validate our method with another well-known mathematical function for GSA studies, the Hartmann 6-Dimensional (6D) function [21]. As the name suggests, this function contains six design variables and is formulated as

$$f(\boldsymbol{x}) = \sum_{i=1}^{4} \alpha_i \exp\left(\sum_{j=1}^{6} A_{ij}\left(x_j - P_{ij}\right)^2\right), for\ \boldsymbol{x} = [x_1, x_2, x_3, x_4, x_5, x_6] \in [0,1], \quad (7)$$

where the variables $\boldsymbol{\alpha}$, $\boldsymbol{A}$, and $\boldsymbol{P}$ are constants and their corresponding values are given in [21]. Implementing a similar transformation as the Ishigami function, we converted two of the design variables, $x_1$ and $x_6$ to qualitative variable, $t_1$ and $t_2$, to create a mixed-variable version as the test case, assigning different discrete values as the levels. The number of discrete options ranged from [2,14] for each variable. The main reason behind choosing these two variables is because their original continuous TSI values, shown in Table 2, contribute to the response the most compared to the remaining four design variables. As a result, the goal is to observe if our method can capture the same contribution. For demonstration purposes, the sensitivity values of the two qualitative variables are shown only. We implemented our method once again through the two-stage validation process.

**TABLE 2.** Ground-truth Sobol' sensitivity indices of two quantitative design variables from the Hartmann 6D function

| Design Variables | Main Sensitivity Index (MSI) | Total Sensitivity Index (TSI) |
|---|---|---|
| $x_2$ | 0.0025 | 0.3992 |
| $x_6$ | 0.0086 | 0.4812 |

Figure 4 shows the well-matching Sobol' sensitivity indices between the true discrete indices (True-MV MSI & TSI) and indices obtained from our method (MV MSI & TSI) for the two qualitative variables. Specifically, the method was able to capture the high contribution of TSI for the two qualitative variables, which was the main goal of this study. Furthermore, the convergence towards ground-truth Sobol' values (True TSI & MSI) demonstrates the effectiveness of our method once again. Although not shown here, the same matching trend is also observed for the remaining four quantitative variables $(x_1, x_2, x_3, x_4)$.

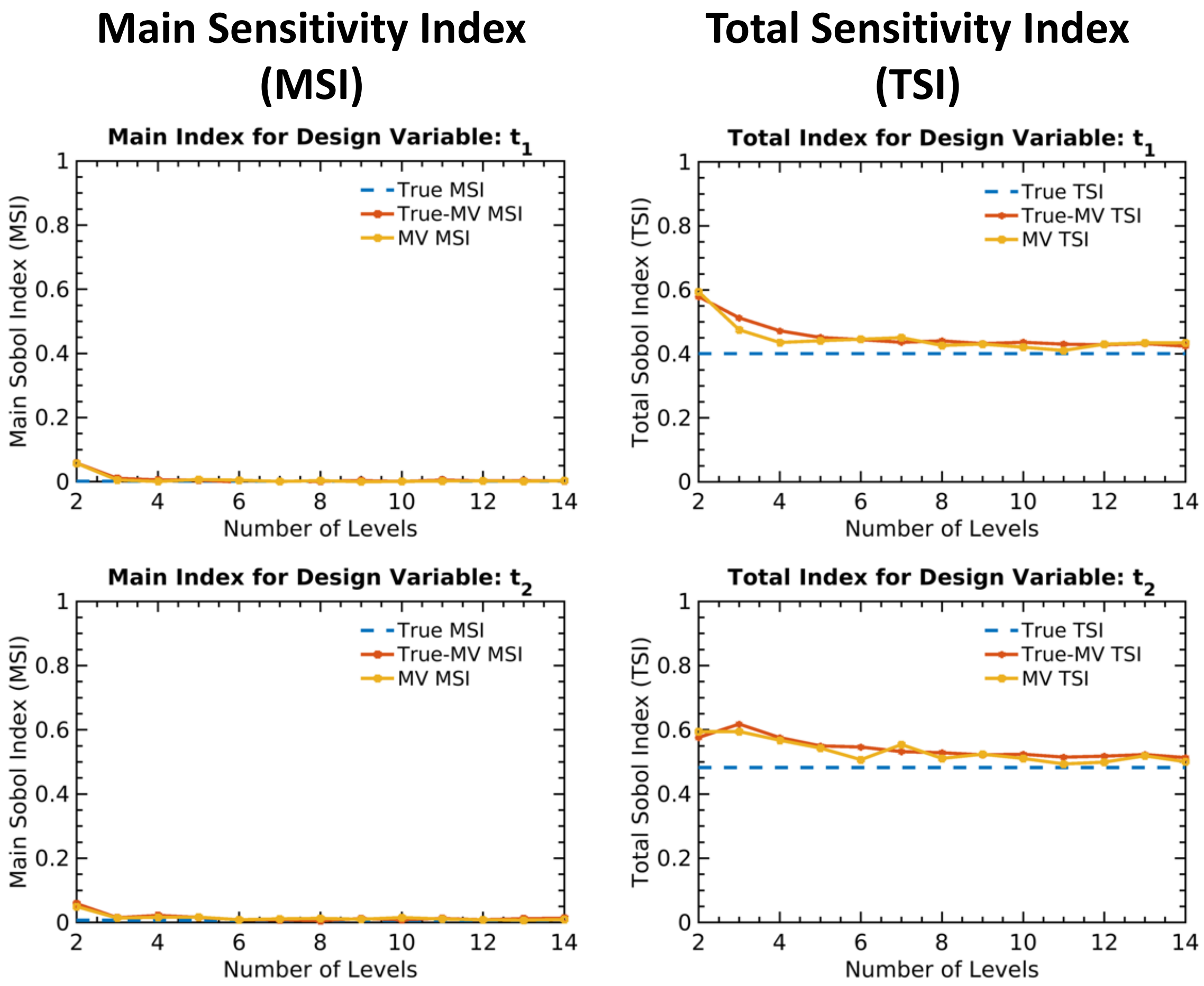

**FIGURE 4**: Sobol' sensitivity indices values of the modified Hartmann 6D function with qualitative inputs

## 4. SENSITIVITY-AWARE BAYESIAN OPTIMIZATION FOR MULTI-OBJECTIVE COMBINATORIAL MATERIALS DESIGN

In this section, we show how our method can be utilized in engineering applications, using the design of metal-organic frameworks (MOFs) as a demonstrative example. Specifically, in Section 4.1 we applied our GSA method to the qualitative design space of MOFs for two gas absorption properties and explained the physical meaning behind the obtained sensitivity indices. Finally, in Section 4.2 we demonstrated how the sensitivity information obtained from our method can benefit the design of MOFs for accelerating the exploration of novel candidates from a large combinatorial design space.

## 4.1. Global Sensitivity Analysis for the Design of Metal-Organic Frameworks for Gas Absorption Properties

MOFs are a class of porous crystalline materials that are formed by arranging organic linkers, inorganic nodes, and functional groups in various topologies. Due to their highly tunable nature, MOFs have been promising solutions for numerous applications including gas storage, gas separation, and catalysis [22-24]. Recently, we have implemented LVGP and Bayesian optimization to identify the Pareto front designs from a large combinatorial design spaces of MOFs for carbon dioxide ($CO_2$) absorption properties [25]. Specifically, we designed MOFs for their carbon capture capabilities, *$CO_2$ working capacity* property, and separation from nitrogen ($N_2$), *$CO_2/N_2$ selectivity* property, as these two properties are involved in many $CO_2$ applications, such as carbon capture from flue gas emissions and natural gas sweetening [26]. In this work, we incorporated GSA into the design optimization to further accelerate the Pareto front exploration of MOFs for the gas absorption properties.

MOFs can be represented by a vector of qualitative variables, [*A*-*B*-*C*-*D*], where each element in the vector takes an integer number to represent the qualitative design choice of that variable. Starting from the element "*A*", each element in the vector corresponds to a qualitative building block design variable, specifically known as Nodular Building Block 1 (N-BB1), Nodular Building Block 2 (N-BB2), Connecting Building Block 1 (C-BB1), and Connecting Building Block 2 (C-BB2), respectively. These variables represent the core building blocks that make up the material. Each choice of building block contains different design options represented by an integer.

In our previous work, we considered a design space of 1001 MOFs constructed from combinations of 7, 4, 6, and 6 unique qualitative design variables of N-BB1, N-BB2, C-BB1, and C-BB2, respectively. The qualitative design variables are shown in Figure 5.

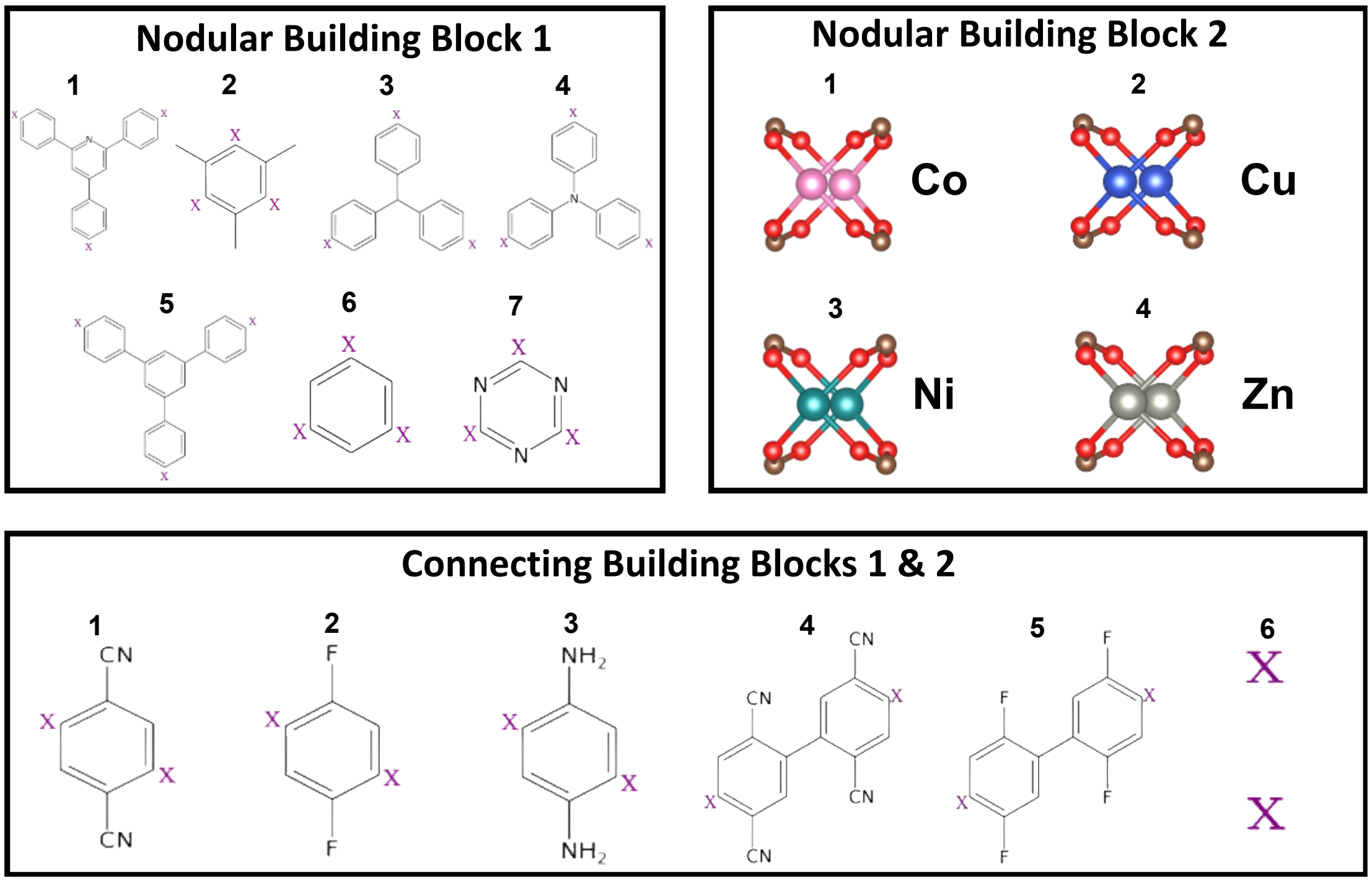

**FIGURE 5**: The design choices for the metal-organic framework (MOF) material construction

Here, we used the same dataset where the two gas absorption properties were obtained through Grand Canonical Monte Carlo (GCMC) simulations for the MOFs. The details of the GCMC simulations are given in our previous paper [25]. Then, we fitted two LVGP models on the entire design space to capture the relationships between the design variables and the two gas absorption properties. Next, we implemented the proposed metamodel-based mixed-variable global sensitivity analysis method to quantify the contribution of qualitative design variables to the variability in the properties. The obtained Sobol' indices, MSI and TSI, for the two properties are displayed in Figure 6.

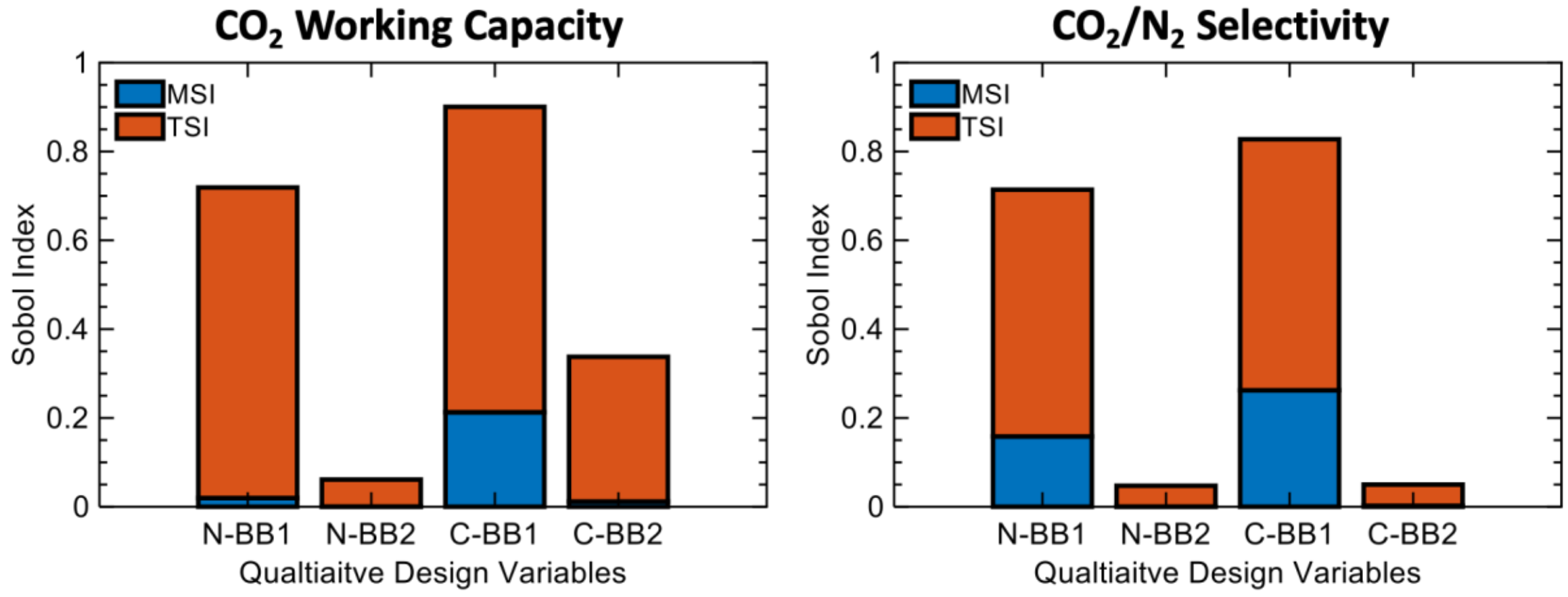

**FIGURE 6**: Sobol' sensitivity indices of the four qualitative design variables on the two properties of interest, $CO_2$ working capacity (left) and $CO_2/N_2$ selectivity (right). Blue and orange bars represent the Main Sensitivity Indices (MSI) and Total Sensitivity Indices (TSI) respectively

The obtained results show that the variables N-BB1 and C-BB1 influence the $CO_2/N_2$ selectivity significantly. Specifically, their interaction effect influences the property as the TSI values of both variables are much higher than others. Similar results are also obtained for $CO_2$ working capacity property. These insights obtained from the GSA method can actually be explained and justified by the chemical mechanism behind the two qualitative variables. Based on the knowledge from material science, the two properties are highly dependent on the size of the MOF structure, which could be defined by the MOF's largest cavity diameter (LCD) [25]. LCD is a microstructural parameter that is defined as the diameter of largest sphere that can be fitted in the pores of the MOFs and provides information about the overall size of the MOF structure. As a result, MOFs with smaller pores could result in stronger van der Waals force interactions which in return can increase gas absorption. As shown in Figure 7, the LCD and both properties are negatively influenced by each other. In other words, MOFs with smaller LCD values tend to have high $CO_2$ gas absorption properties and vice versa. Meanwhile, the LCD of the MOF is controlled by the building blocks that surround the material. Therefore, we can use LCD as an intermediate descriptor to unravel the relationship between the structural design variables and properties to justify the proposed mixed-variable GSA.

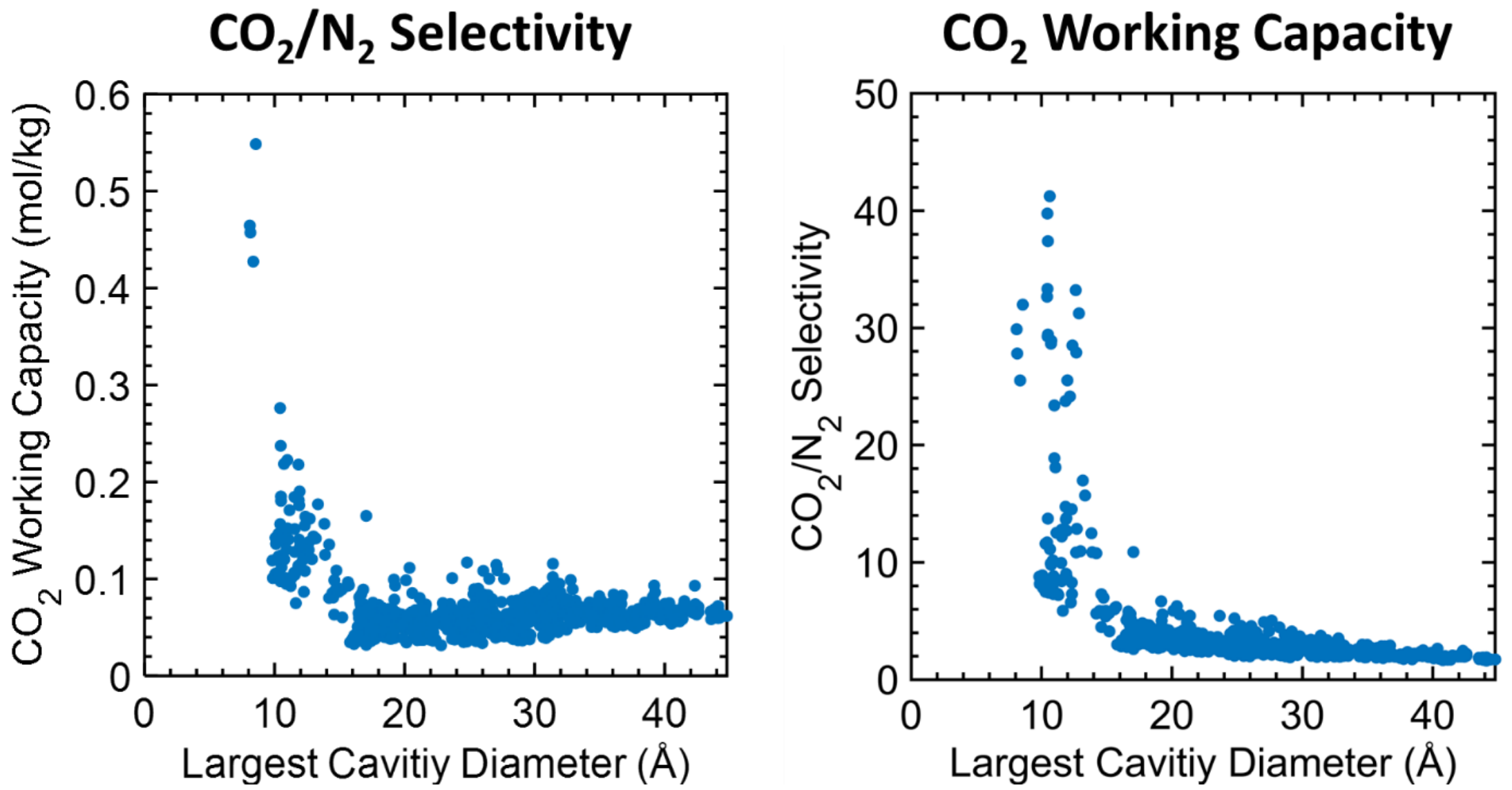

**FIGURE 7**: Largest cavity diameter (LCD) compared with the two $CO_2$ gas absorption properties, $CO_2$ working capacity (left) and $CO_2/N_2$ selectivity (right)

Specifically, the N-BB1 and C-BB1 determine the overall shape and size of the material. As shown in Figure 5, the N-BB2 is a choice of metal node, which does not influence the shape significantly. Similarly, the C-BB2 connects the N-BB2 to the MOF, which does not contribute to the overall shape of the structure. Figure 8 further shows the variations of LCD with respect to design choices within each design variable. It is immediately apparent that the choice design variables for N-BB1 and C-BB1 significantly change the LCD. In contrast, the choice of N-BB2 or C-BB2, does not have much influence on the LCD. Thus, the LCD distributions of the variables can explain the sensitivity values obtained from our method.

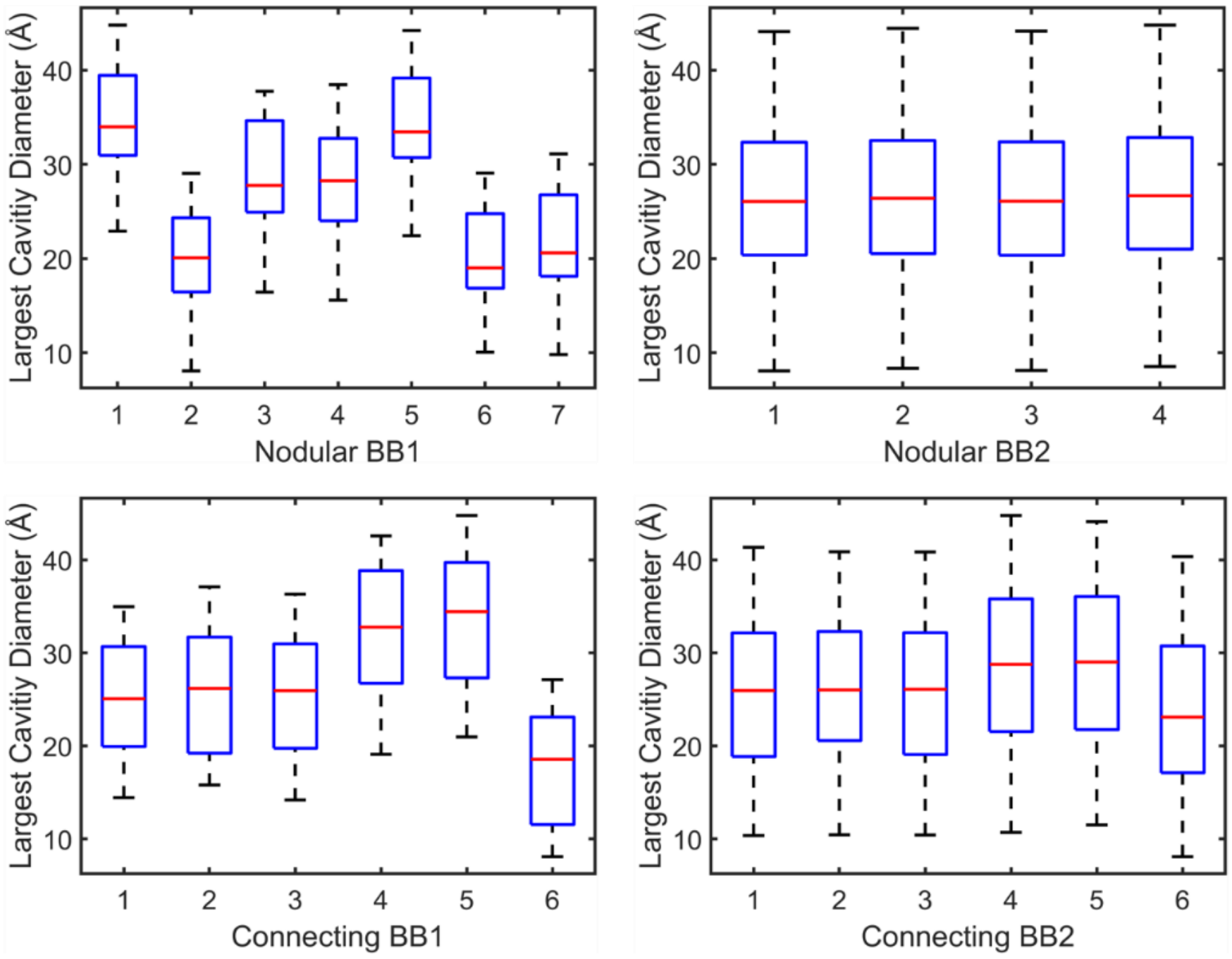

**FIGURE 8**: Largest cavity diameter (LCD) distribution with respect to different building block design choices. For each box, the red line indicates the median, the bottom and top edges of the box indicate the 25th and 75th percentiles, and the whiskers at the top and bottom represent the maximum and minimum observed LCD values of the specific design variables (level), respectively

The qualitative GSA enabled us to conclude that Nodular BB1 and Connecting BB2 are the design variables that contribute to the two absorption properties the most. In the next section, we used this important knowledge to accelerate the discovery and design of novel MOFs.

## 4.2. The Sensitivity Aware Multi-Objective Design

### 4.2.1. The Design Framework

In this section, we implemented a new sensitivity-aware design framework using the knowledge obtained from the GSA study for the design of MOFs for the two gas absorption properties, $CO_2$ working capacity and $CO_2/N_2$ selectivity. Our design optimization goal is to identify MOF designs that lie on the Pareto front of the two conflicting gas absorption properties. Here, the Pareto front corresponds to MOF designs that possess properties higher than the rest of the design space but cannot be improved without sacrificing the other property of interest [27]. We will first explain the new sensitivity-aware framework, then demonstrate its results, and finally compare the new framework with the "vanilla" design framework without using the sensitivity analysis [25].

The new sensitivity-aware Bayesian optimization design framework consists of two stages as shown in Algorithm 1. For ease of notation, the four design variables are represented using the vector notation, [*A, B, C, D*] and their respective two properties, $CO_2$ working capacity ($Y_1$) and $CO_2/N_2$ selectivity ($Y_2$), are described with [$Y$] = [$Y_1$, $Y_2$]. In the first stage, the framework searches for optimum design candidates only in the Nodular BB1 (*A*) and Connecting BB1 (*C*) space. This is because, based on the conclusions obtained from our GSA framework, they are the most important factors for gas absorption properties, while the remaining two factors have little influence. By focusing only on those important factors, we can significantly reduce the search space to facilitate design optimization.

**ALGORITHM 1:** Sensitivity-Aware Bayesian Optimization

*Start* First Stage
$D_{ABCD} \leftarrow [(A, B, C, D), Y]$
**for** *iteration* = 1, 2, . . . , *n* **do**
    Train $LVGP_{ABCD}$ with $D_{ABCD}$
    $D_{AC} \leftarrow [(A, C), \max_{(B,D)} (\widehat{Y}_{ABCD} \mid A, C)]$
    Train $LVGP_{AC}$ with $D_{AC}$
    Find $A_{can}$, $C_{can}$ with BO and $LVGP_{AC}$
    **for** each $B$, $D$ given $A_{can}$, $C_{can}$ **do**
        Find $B_{can}$, $D_{can}$ with BO & $LVGP_{ABCD}$
    **end for**
    Add $[(A_{can}, B_{can}, C_{can}, D_{can}), Y_{can}]$ to $D_{ABCD}$
**end for**
Obtain $(A_{opt}, C_{opt})$ from $D_{AC}$
*End* First Stage
*Start* Second Stage
**while** $(A_{PF}, B_{PF}, C_{PF}, D_{PF}) \notin D_{ABCD}$ **do**
    **for** each $X_B$, $X_D$ given $(A_{opt}, C_{opt})$ **do**
        Find $A_{can}$, $B_{can}$, $C_{can}$, $D_{can}$ with BO & $LVGP_{ABCD}$
    **end for**
Add $[(A_{can}, B_{can}, C_{can}, D_{can}), Y_{can}] \rightarrow D_{ABCD}$
Train $LVGP_{ABCD}$ with $D_{ABCD}$
**end while**
Obtain $(A_{PF}, B_{PF}, C_{PF}, D_{PF})$ from $D_{ABCD}$
*End* Second Stage

To initiate the first stage, a design of experiment (DOE) with all four variables is created from sliced-optimum Latin hypercube sampling (Sliced-OLHS) [28]. In order to facilitate notation, this initial design space, along with its corresponding properties, is denoted as $D_{ABCD}$. Two LVGP models are trained on the $D_{ABCD}$ for each property with all four design variables as inputs, denoted as $LVGP_{\mathrm{ABCD}}$. To simplify notation, the $LVGP_{ABCD}$ contains two LVGP models trained on the two properties. We use these two models to make predictions on all the MOF design candidates for each property. The predicted properties for both objectives are represented with $\hat{Y}_{ABCD}$. Next, we create a new dataset, $\mathrm{D_{AC}}$, from the initial DOE with only the unique [*A*, *C*] combinations as the input. For each unique [*A*, *C*], we collect the maximum property values predicted by the $LVGP_{ABCD}$ models for all combinations of [*B*, *D*] as the outputs for the new dataset. In other words, the outputs represent the best properties we are expected to get for a given [*A*, *C*]. A second set of two LVGP models, $LVGP_{AC}$, is then trained on this new dataset with only unique [*A*, *C*] combinations as the input and their respective predicted maximum property values as outputs.

A multi-objective Bayesian optimization (BO) is then implemented in this first stage to search for the most optimal [*A*, *C*] combination using $\mathrm{LVGP_{AC}}$ to obtain the acquisition function, i.e., Expected Improvement (EI) [29]. For the multi-objective scenarios of this study, BO aims to identify the Pareto front solutions. This multi-objective BO identifies the [*A*, *C*] candidate, denoted as [$A_{can}$, $C_{can}$] that is expected to improve and expand the current known Pareto front designs in the [*A*, *C*] space the most. However, a full MOF representation vector including both Nodular BB2 (*B*) and Connecting BB2 (*C*) is needed to evaluate the true gas absorption values. To mitigate this, a second multi-objective BO is run with fixed [$A_{can}$, $C_{can}$] to identify the best [*B*, *D*] candidate, denoted as [$B_{can}$, $D_{can}$], using $LVGP_{ABCD}$ to obtain the EI values. After that, a new MOF candidate with full representation, i.e., [$A_{can}$, $B_{can}$ $C_{can}$, $D_{can}$], is identified and passed to property evaluations to complete one iteration of the first stage. This candidate is added to the list of explored MOF candidates and the $LVGP_{ABCD}$ model is updated accordingly for the next iterations of the first stage of the framework. The first stage is terminated after reaching the predefined maximum number of iterations, with all the optimum candidates [$A_{opt}$, $C_{opt}$] on the Pareto front (PF) identified.

In the second stage, the original design space has been reduced to a much smaller space, in which [$A$, $C$] can only be selected from the optimized [$A_{opt}$, $C_{opt}$] in the first stage. Again, we use BO to search for the Pareto front set in this reduced design space with $LVGP_{ABCD}$ as the surrogate model. At the end of the second stage, the optimum candidates representing the Pareto front [$A_{PF}$, $B_{PF}$, $C_{PF}$, $D_{PF}$] are evaluated and the design process is concluded.

The GSA results obtained from our framework enabled us to direct the Initial focus of the exploration toward novel Nodular BB1 and Connecting BB2 candidates.

#### 4.2.2. Design Results

In this section, we demonstrate the capability of our sensitivity-aware design framework on the MOF design space and compare it with a "vanilla" design framework. Without the GSA information, the vanilla framework directly uses four-variable LVGP models in the multi-objective BO to search for the Pareto front in the original full design space. This large MOF design space contains 47,470 MOF candidates constructed from combinations of 4, 7, 41, and 42 combinations of Nodular Building Block 1 (N-BB1), Nodular Building Block 2 (N-BB2), Connecting Building Block 1 (C-BB1), and Connecting Building Block 2 (C-BB2), respectively. We have previously simulated the properties of all 47,740 candidates using GCMC simulations and identified 7 Pareto Front candidates to be explored by both frameworks. The details of the GCMC simulations can be found in [25]. The property space, along with the Pareto front points of two properties is given in Figure 9, which will be used as the ground truth for later validation.

To account for at least one of the design options (levels) of the C-BB2, the vanilla framework started optimization with a DOE that contains 42 MOFs obtained from Sliced-OLHS. After continuing for 190 iterations of multi-objective BO, the vanilla framework identified all the Pareto front designs, exploring 232 MOFs in total. The number of explored MOFs corresponded to 0.49% of the entire design space, which demonstrates the effectiveness of LVGP on large combinatorial design spaces once again. This result was used as a benchmark for our sensitivity-aware design framework.

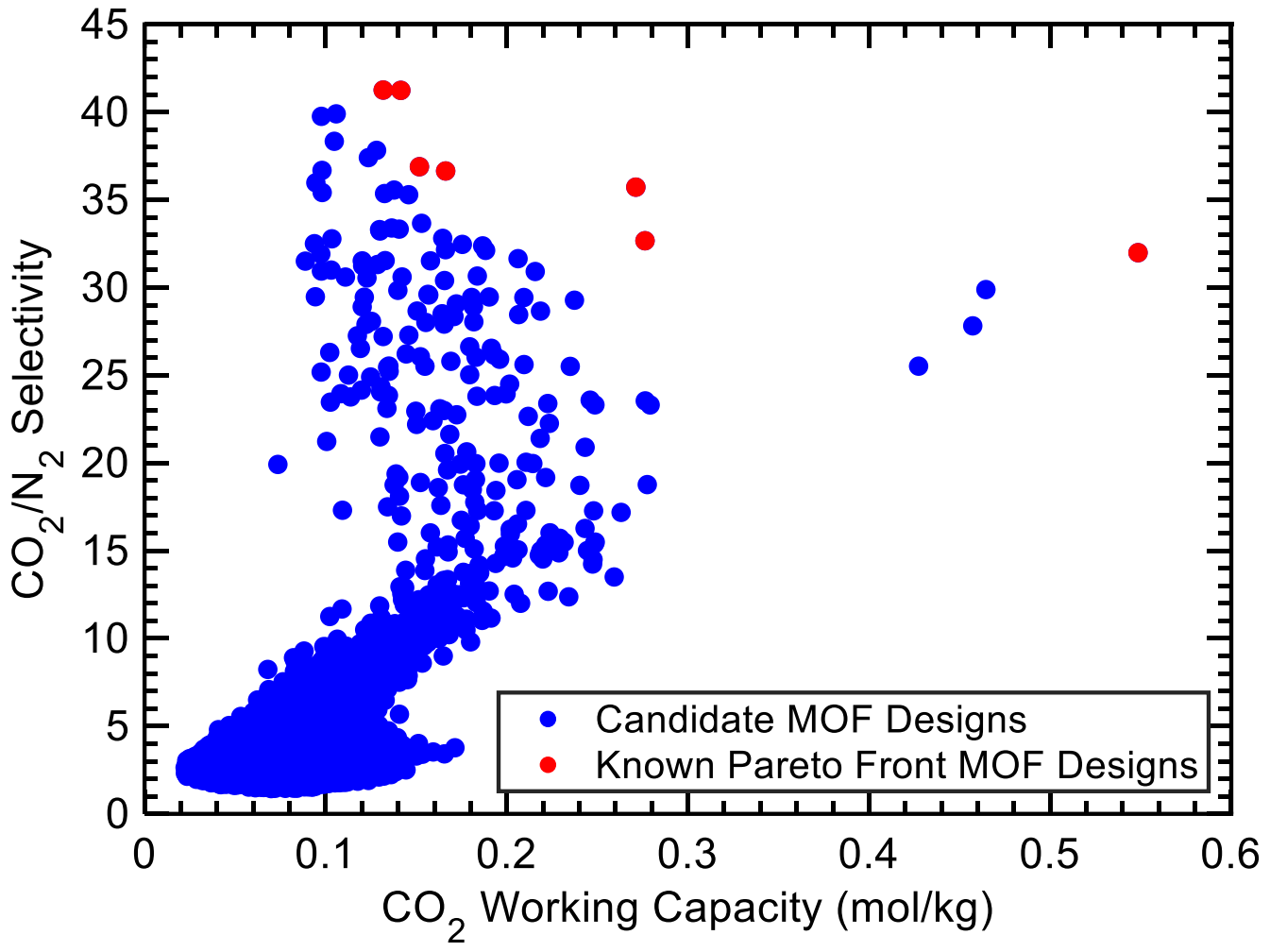

**FIGURE 9**: The known property space of $CO_2$ working capacity and $CO_2/N_2$ selectivity properties. Red and blue points indicate the known Pareto front and the remaining properties, respectively

Starting with the same DOE used in the vanilla framework, we ran the first stage of the sensitivity-aware design framework until 100 MOFs are explored, including the 42 initial MOFs, to identify the optimal [$A_{opt}$, $C_{opt}$] design variables. At the end of this 58-iteration optimization, two optimal [$A_{opt}$, $C_{opt}$] candidates are identified and passed on to the second stage. The optimal candidates contained two different design options for the Nodular BB1 (*A*) and only one design variable for the Connecting BB1 (*C*). In the second stage, the third BO framework is run to identify the optimal candidates from a much smaller design space, as the design space searched reduced from 47,740 candidates to a design space of 336 candidates that are formed by combinations of 2 N-BB1, 4 N-BB2, 1 C-BB1, and 42 C-BB2. Using the previous $LVGP_{ABCD}$ model, the second stage is initiated to identify the true Pareto Front designs, [$A_{PF}$, $B_{PF}$, $C_{PF}$, $D_{PF}$]. After 113 iterations of the second stage, all the Pareto front MOF designs were identified. Including the 100 MOFs identified from the first stage, a total of 213 MOFs were explored to find the true Pareto front MOF designs, corresponding to 0.45% of the design space. Figure 10 and Figure 11 show the optimization history search and the evolution of the Pareto front for both frameworks, respectively. As shown in Figure 10 (a), the sensitivity-aware design framework was able to identify all Pareto designs at a faster rate compared to the vanilla framework. Similarly, the Pareto evolution shown in Figure 10 (b) demonstrates that the sensitivity-aware design framework can swiftly expand the Pareto front. Here, the colors represent the optimization iteration history where lighter and darker colors indicate the earlier and later stages of optimization, respectively. Specifically, we can observe lighter color points with a much higher property values identified by the sensitivity-aware design framework, while the vanilla framework achieves similar Pareto front properties at the later stages of optimization which is denoted by darker points. This result shows that the sensitivity-aware design framework can expand the Pareto front at a faster rate compared to the vanilla framework, particularly during the earlier stages of design optimization. This can be very advantageous for optimization applications with limited resources.

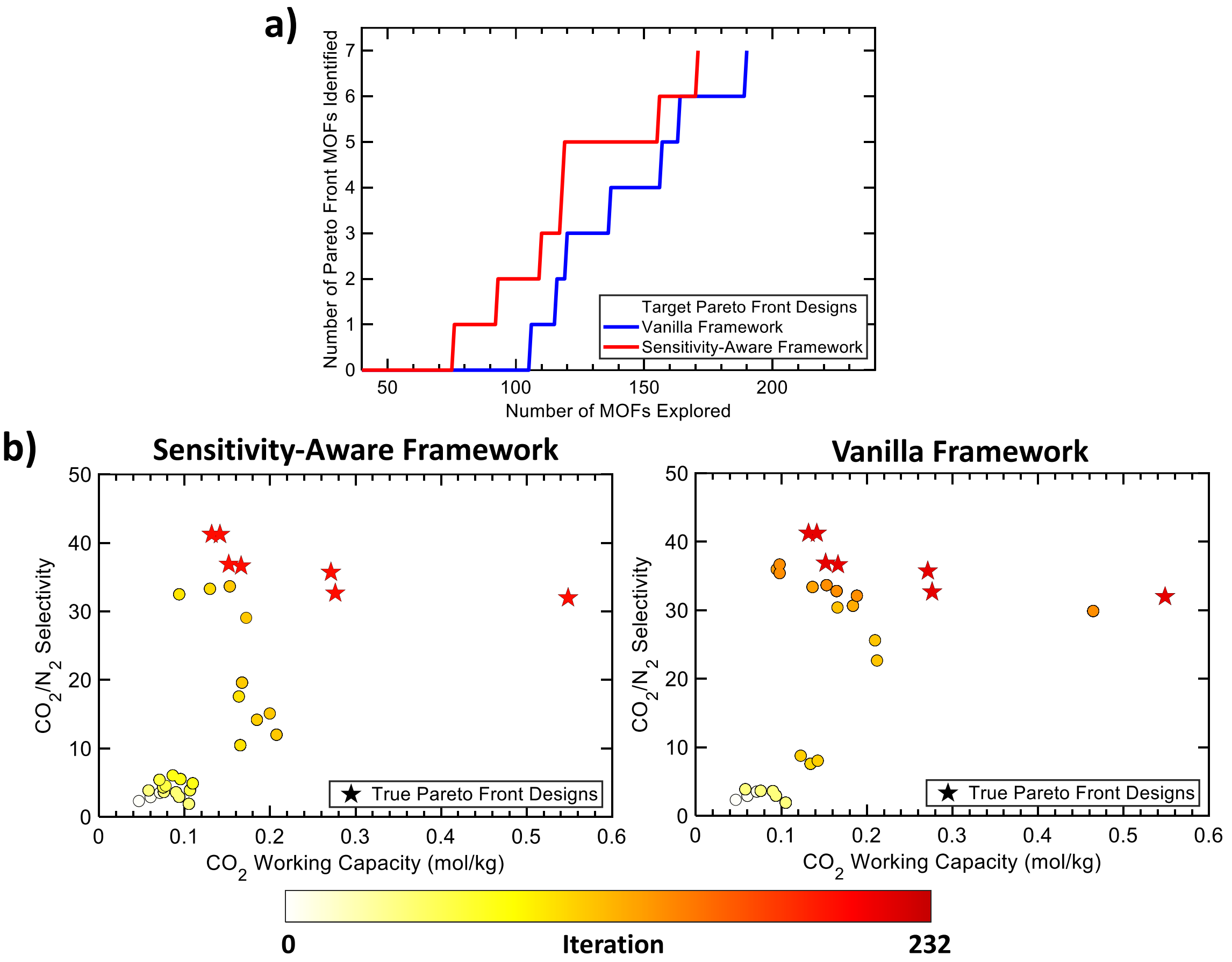

**FIGURE 10**: Pareto front optimization (a) and evolution (b) history comparison between the sensitivity-aware and vanilla design frameworks

Moreover, compared with the vanilla BO, the sensitivity-aware design framework reduces the number of MOFs explored by 8.2%, using only 0.44% of the whole design space to successfully identify the Pareto front. Since the evaluation of properties is relatively expensive (approximately 4 hours per MOF), this 8% improvement provides notable time and cost savings for designers. It can be expected that the level of improvement will become higher for other design applications with more variables and larger design spaces.

The obtained results show that our metamodel-based mixed-variable GSA method can play an important role in the design of MOFs. The method not only provides physical insights regarding the design choices but also can assist in better design optimization by directing the focus of optimization towards better solutions to significantly reduce the search space.

## 5. CONCLUSION

Sensitivity analysis is a very powerful technique to evaluate design models in multiple aspects. However, current studies under global sensitivity analysis techniques fail to consider mixed-variable design spaces. In this paper, we developed a metamodel-based global sensitivity analysis method that can accommodate models with mixed variables or only qualitative variables. We validated the effectiveness of our method on two well-known mathematical equations used in sensitivity studies. This validation concluded that our method is able to capture the relationships between and within mixed-variable design spaces on the response of interest effectively. We demonstrated the benefit of incorporating our method into combinatorial and many-level engineering materials design applications. Specifically, the newly developed sensitivity-aware design framework greatly accelerated the exploration of novel material candidates, providing superior efficiency in the design of metal-organic framework materials for gas absorption applications. We envision that our new metamodel-based mixed-variable global sensitivity analysis method can further assist design engineers in building more accurate mixed-variable design models by identifying the most influential design variables, observing input and output uncertainties, and calibrating model parameters.

## ACKNOWLEDGMENT

We acknowledge the support from Air Force Office of Scientific Research (AFOSR) under award number AFOSRFA9550-18-1-0381. Finally, we acknowledge Professor Randall Q. Snurr, and Thang Duc Pham from Department of Chemical and Biological Engineering at Northwestern University for generating the metal-organic framework data, and Yi-Ping Chen from the Department of Mechanical Engineering at Northwestern University for helping with the manuscript figures.

## Figure Captions List

Figure 1. The mapping from qualitative design variables onto 2-dimensional quantitative latent space.

Figure 2. The flowchart of the metamodel-based mixed-variable global sensitivity analysis method.

Figure 3. Sobol' sensitivity indices of the modified Ishigami function with mixed-variable inputs.

Figure 4. Sobol' sensitivity indices values of the modified Hartmann 6D function with qualitative inputs.

Figure 5. The design choices for the metal-organic framework (MOF) material construction.

Figure 6. Sobol' sensitivity indices of the four qualitative design variables on the two properties of interest, $CO_2$ working capacity (left) and $CO_2/N_2$ selectivity (right). Blue and orange bars represent the Main Sensitivity Indices (MSI) and Total Sensitivity Indices (TSI) respectively.

Figure 7. Largest cavity diameter (LCD) compared with the two $CO_2$ gas absorption properties, $CO_2$ working capacity (left) and $CO_2/N_2$ selectivity (right).

Figure 8. Largest cavity diameter (LCD) distribution with respect to different building block design choices. For each box, the red line indicates the median, the bottom and top edges of the box indicate the 25th and 75th percentiles, and the whiskers at the top and bottom represent the maximum and minimum observed LCD values, respectively.

Figure 9. The known property space of $CO_2$ working capacity and $CO_2/N_2$ selectivity properties. Red and blue points indicate the known Pareto front and the remaining properties, respectively.

Figure 10. Pareto front optimization (a) and evolution (b) history comparison between the sensitivity-aware and vanilla design frameworks.

## Table Caption List

Table 1. Ground-truth Sobol' sensitivity indices of the original Ishigami function with only quantitative inputs.

Table 2. Ground-truth Sobol' sensitivity indices of two quantitative design variables from the Hartmann 6D function.